\documentclass[conference]{IEEEtran}
\IEEEoverridecommandlockouts
\usepackage{amsmath,amssymb,amsfonts}
\usepackage{graphicx}
\usepackage{subfigure}
\usepackage{booktabs}
\usepackage{marvosym}
\usepackage[colorlinks,
            linkcolor=blue,
            anchorcolor=blue,
            citecolor=blue]{hyperref}
\def\BibTeX{{\rm B\kern-.05em{\sc i\kern-.025em b}\kern-.08em
    T\kern-.1667em\lower.7ex\hbox{E}\kern-.125emX}}
\begin{document}

\title{Medical Speech Symptoms Classification via Disentangled Representation\\
}

\author{
\IEEEauthorblockN{
Jianzong Wang$^{1\dag}$, 
Pengcheng Li$^{1,2\dag}$ \thanks{\dag \, Equal contribution.}, 
Xulong Zhang$^{1\textsuperscript{\Letter}}$ \thanks{\textsuperscript{\Letter} Corresponding author.},  
Ning Cheng$^{1}$, 
Jing Xiao$^{1}$}
\IEEEauthorblockA{
\textit{$^{1}$Ping An Technology (Shenzhen) Co., Ltd.} \\
\textit{$^{2}$University of Science and Technology of China}}
\IEEEauthorblockA{
\texttt{jzwang@188.com, pechola.lee@outlook.com,
zhangxulong@ieee.org, }
\\ \texttt{\{chengning211, xiaojing661\}@pingan.com.cn
}}}

\maketitle

\begin{abstract}
\textit{Intent} is defined for understanding spoken language in existing works. Both textual features and acoustic features involved in medical speech contain intent, which is important for symptomatic diagnosis. In this paper, we propose a medical speech classification model named \textbf{DRSC} that automatically learns to disentangle intent and content representations from textual-acoustic data for classification. The intent representations of the text domain and the Mel-spectrogram domain are extracted via intent encoders, and then the reconstructed text feature and the Mel-spectrogram feature are obtained through two exchanges. After combining the intent from two domains into a joint representation, the integrated intent representation is fed into a decision layer for classification. Experimental results show that our model obtains an average accuracy rate of 95\% in detecting 25 different medical symptoms.
\end{abstract}

\begin{IEEEkeywords}
    medical speech, multi-modal neural network, speech representation disentanglement
\end{IEEEkeywords}

\section{Introduction}
The questioning ability of medical students needs to be accumulated for a long time \cite{daniel2019clinical}, however, it is difficult for schools to allow students to study for decades to achieve a high level of questioning ability. The current solution is to recognize intent from doctor-patient interviews \cite{rojowiec2020intent}, by annotating the intent of the utterances in advance. Blackley \textit{et al.} \cite{blackley2019speech} point out that speech recognition for clinic documentation is increasingly common, but is also heterogeneous. Narayanan \textit{et al.} \cite{narayanan2004transonics} propose a speech system that focuses on speech recognition, translation and dialogue management, aiming to assist in medical communication.

With the development of deep learning in recent years, various intelligent medical signal processing or healthcare systems based on deep learning are proposed \cite{muhammad2021comprehensive}. The general idea behind these diagnosing symptoms is to convert speech to text via automatic speech recognition (ASR) models, and then perform sentence grammar analysis on the text. 

However, human speech contains not only textual information (\textit{i.e.} linguistic information) but also a variety of acoustic information  (\textit{e.g.} pitch, rhythm and emotion) related to the intent \cite{gu2017speechic}. But most early deep learning works ignore the acoustic feature of patients' speeches.  SpeechIC \cite{gu2017speechic} is a intent classification model which extracts intent from text and audio domains. They design a model that uses both textual features and acoustic features for intent detection, and learns the embedding from textual and acoustic features through a convolutional neural network (CNN), then directly fuses the features of the two domains. But the success of these methods is due to the powerful feature extraction capabilities of CNN, which does not specifically analyze intent features related to medical symptoms speech. The extraction of intent for classification faces challenges, as the efficiency of the extraction process can only be restricted by the classification results. 

In order to effectively disentangle intent representations from medical speech and improve the accuracy of medical speech symptoms classification, we propose a model named \textbf{DRSC} which disentangles intent information from both text and Mel-spectrogram domains. The multi-modal model puts textual and acoustic features into the generative network separately for disentanglement, the disentangled representations are then fused and sent to a classifier for medical symptoms classification. In this way, information related to symptoms is extracted from both textual and acoustic features of speech and then contributes to the diagnosis of symptoms. Experimental results show that DRSC achieves satisfactory performance, and the experiments on inaccurate transcriptions show that our model owns robustness. Our main contributions can be summarized as follows:
\begin{itemize}
\item We introduce a multi-modal representation disentangling framework that factorizes domain-specific content features and domain-invariant symptom intent features.
\item Our proposed model achieves competitive performance in terms of accuracy and robustness on the Medical Speech, Transcription, Intent dataset.
\end{itemize}

\begin{figure*}[!ht]
\centering
\includegraphics[width=17cm]{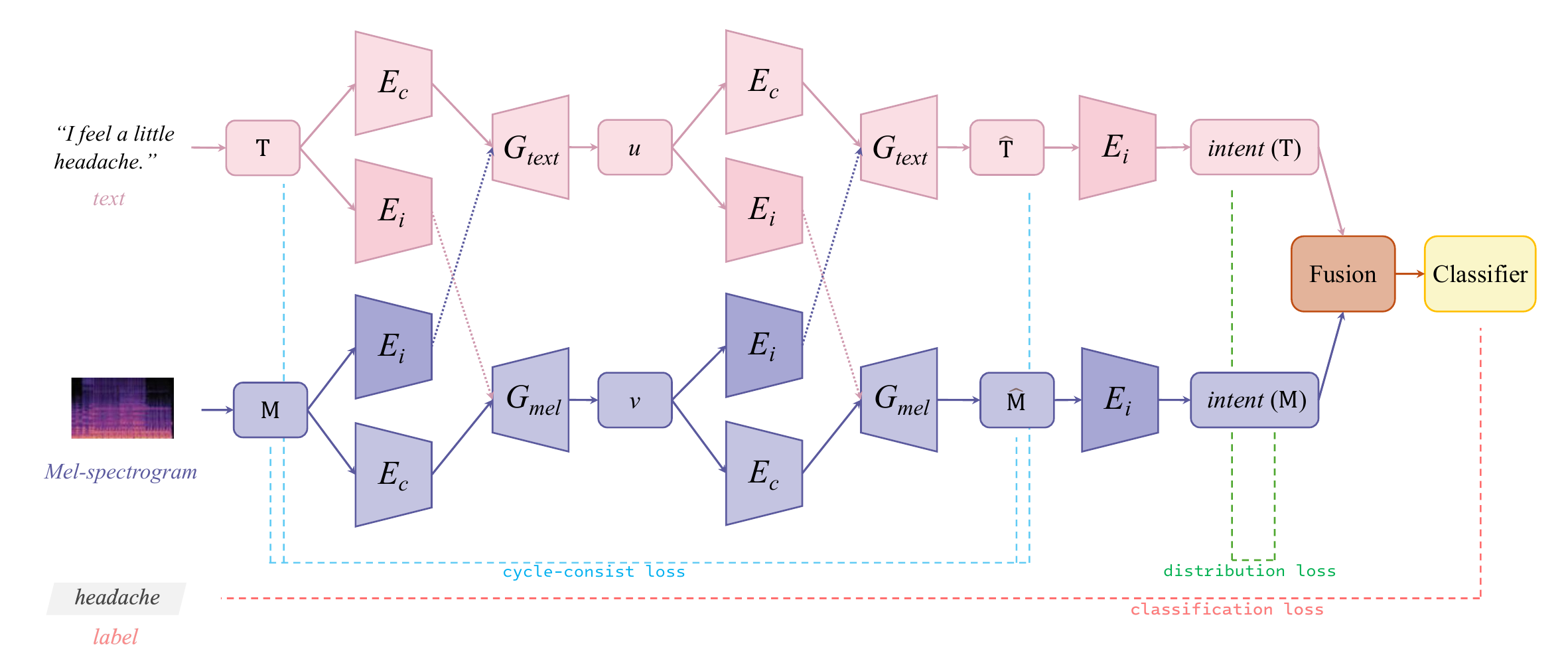}
\caption{Architecture and basic training objectives of \textbf{DRSC}, the model disentangles intent representations from text and Mel-spectrogram domains. Intent information extracted from the two domains is then fused for classification.}
\label{fig:framework}
\end{figure*}

\section{Related Work}
\subsection{Medical Speech Symptoms Classification}
A common solution for medical speech symptoms classification is to conduct feature engineering in the textual domain and combine neural networks for classification \cite{kyle2021}. However, this series of methods only consider text but ignore the original acoustic information of speech. Because speech with the same text content may carry different intent information due to the speaking condition, these acoustic features are essential for understanding the underlying problems of patients. Previous studies \cite{yoon2018multimodal,christy2020multimodal} show that the combination of textual features and acoustic features achieves good performance in speech emotion classification. With the development of deep learning, Mittal \textit{et al.} \cite{mittal2020m3er} propose a multi-modal model named M3ER that extracts cues from face, text and speech for emotion recognition, and its test conducted on multi-modal datasets achieves good results.

\subsection{Generative Adversarial Network}
Generative Adversarial Network (GAN) \cite{ian2014gan} consists of a generator $G$ and a discriminator $D$ and its training approach is based on a game theory. In order to achieve the best performance, $D$ is trained to distinguish a real sample $x$ from a synthetic sample $G(z)$ created by $G$ with random noise $z$, and $G$ is trained to synthesize samples as realistic as possible that can cheat $D$. In general, the adversarial relation between $G$ and $D$ is formulated in Eq. \ref{eq:1}:
\begin{equation}
    \label{eq:1}
\begin{array}{r}
\mathop {\min }\limits_G \mathop {\max }\limits_D V(D,G) = {E_{x \sim p(x)}}[\log D(x)] + \\
{E_{z \sim p(z)}}[\log (1 - D(G(z)))]
\end{array}
\end{equation}
where $E$ represents the expected value and $z$ follows sample distribution $p(z)$. $D$ is trained to minimize $\log D(x)$ while $G$ is trained to minimize $\log (1-D(G(z)))$ (\textit{i.e.} maximize $log(D(G(z)))$). In other words, $G$ and $D$ are optimized by the following formulas respectively:
\begin{equation}
\begin{array}{r}
\mathop {\max }\limits_D {V_D}(D,G) = {E_{x \sim p(x)}}[\log (D(x)] + \\
{E_{z \sim p(z)}}[\log (1 - D(G(z)))]
\end{array}
\end{equation}
\begin{equation}
\mathop {\max }\limits_G {V_G}(D,G) = {E_{z \sim p(z)}}[\log (D(G(x))]
\end{equation}

Research on GAN can be divided into two categories. One is to enhance the stability of the training \cite{shahbazi2021collapse}, and the other is to apply GAN to specific tasks \cite{bai2018sod,singh2021medical}. In this paper, we leverage GAN to disentangle speech representations for medical speech symptoms classification.

\subsection{Representation Disentanglement}
Learning the disentangled representations of interpretable generative factors of data is one of the foundations to allow artificial intelligence to think like people. Many types of research achieve successful representation disentanglement of image and speech for various low-level tasks, such as image translation \cite{lee2018diverse} and voice conversion \cite{yang2022speech}. Mathieu \textit{et al.} \cite{mathieu2016disen} combine conditional generative adversarial network and variational autoencoder for disentanglement. Odena \textit{et al.} \cite{odena2017conditional} introduce an auxiliary classifier, which can achieve class-related representation disentanglement. Lee \textit{et al.} \cite{lee2018diverse}  introduces a novel cross-cycle consistency loss to embed images into two spaces for representation disentanglement, and DRVC \cite{wang2022drvc} leverages this framework for voice conversion.

\section{Method}
The front-end framework of DRSC is inspired by an image style transferring model, DRIT \cite{lee2018diverse}. Considering our multi-domain classification task, the key point to our method is the disentanglement of intent representations related to medical speech classification task from both text and Mel-spectrogram domains, and the disentangled information must be highly relevant to this medical task, in order to achieve efficient medical speech symptoms classification. We are going to introduce the proposed model and analyze its modules as well as loss functions.

\subsection{GAN-based Disentanglement}
As shown in Fig. \ref{fig:framework}, DRSC leverages GAN for speech representation disentanglement. In order to extract features highly correlated to the true intent, we use an operation called \textit{cross and restore} for representation disentanglement and jointly use three designed loss functions to ensure the performance of disentanglement.

Our front-end framework consists of four groups of content encoders, intent encoders and generators. The network takes textual feature $T$ (\textit{i.e.} word vectors) and acoustic feature $M$ (\textit{i.e.} Mel-spectrogram) as inputs. Then disentangle the content and intent representations through the content encoder and intent encoder respectively. Firstly, exchange the intent representations between the text domain and the Mel-spectrogram domain, (\textit{i.e.} the content representation from the text domain and the intent representation from the Mel-spectrogram domain are fed into the generator to generator $G_{text}$ to obtain $u$, while the content representation from the Mel-spectrogram domain and the intent representation from the text domain are fed to generate $v$), then repeat the representation disentanglement and exchange the intent representations from two domains (\textit{i.e.} intent representations extracted from $u$ and $v$) for the second time, as well as feeding the content representations to the generators respectively. Finally, the text features $\hat{T}$ and Mel-spectrogram features $\hat{M}$ are reconstructed.

We embed the text domain and Mel-spectrogram domain encoding into \textit{shared intent space} $ {Z_i} $ and \textit{domain-specific content spaces} ${Z_{c_T}, Z_{c_M}}$, that is to say, the intent encoders disentangle intent representations from two domains into $ {Z_i} $, while the content encoders disentangle the remaining different content representations from two domains into $Z_{c_T}$ and $Z_{c_M}$, as shown in Eq. \ref{eq:domain}. Our proposed method is based on an assumption that must be established: common intent information exists in both the text domain and the Mel-spectrogram domain, and it is indistinguishable.
\begin{equation}
\begin{array}{l}
\{ z_{text}^c, z_{text}^i\}  = \{ {E^c_{text}}(T), {E^i_{text}}(T)\} \\
\{ z_{mel}^c, z_{mel}^i\}  = \{ {E^c_{mel}}(M), {E^i_{mel}}(M)\} 
\end{array}
\label{eq:domain}
\end{equation}
where $ z_{text}^c $ is the content representation disentangled from the text domain, $ z_{text}^i $ is the intent representation in the text domain, and the same goes for $ z_{mel}^c $ and $ z_{mel}^i $.

\begin{figure}
\centering
\includegraphics[width=0.5\textwidth]{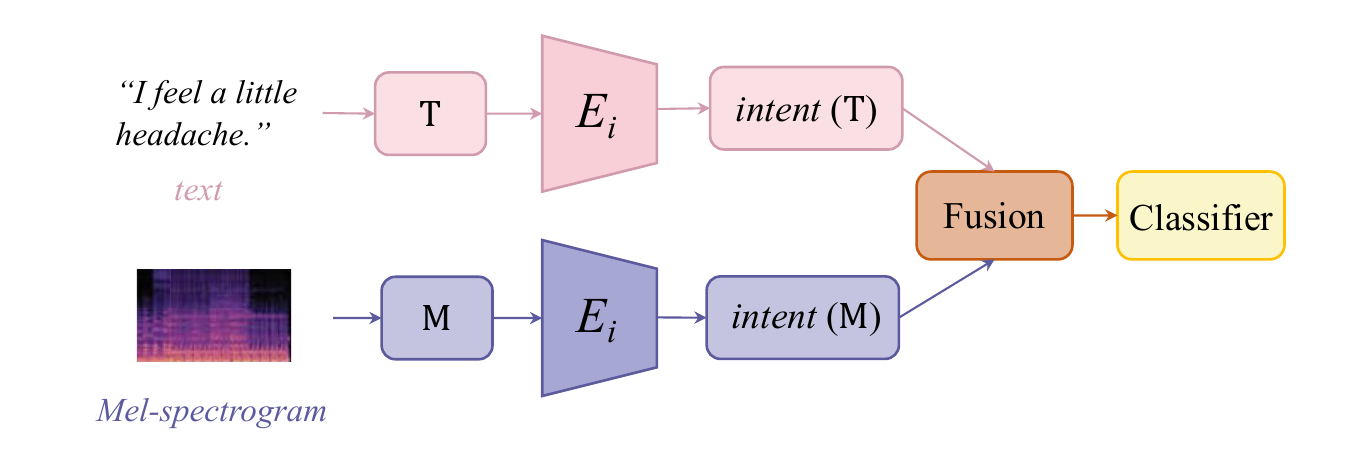}
\caption{In the inference stage, we only need to use the trained intent encoder to extract intent representations from two domains for classification.}
\label{fig:test}
\end{figure}

During the inference stage, as shown in Fig. \ref{fig:test}, we only need to use the parameter-trained intent encoder to disentangle the intent representations from the given text and Mel-spectrogram, and then perform feature fusion on these two extracted intent representations in the intent space, finally put the fused representation into the softmax layer for classification.

The content encoder and intent encoder in DRSC are composed of a convolution bank which contains various kernels with different sizes, as well as multiple \texttt{Conv1d} blocks with residual connections. The intent encoder outputs intent representation through a dense block after convolution. The generator is also composed of several \texttt{Conv1d} blocks with residual connections.

\subsection{Feature Fusion Layer}
Since DRSC conducts classification according to multi-modal information, the fusion of intent information is necessary for its decision. We introduce feature-level fusion for DRSC as the representations from two domains disentangled via DRSC are both in the same space $Z_i$, and can directly perform feature-level fusion. 
The feature fusion layer is simply composed of fully connected layers, which fuse intent representations extracted from the text and Mel-spectrogram into a joint representation.

\subsection{Loss Functions}
DRSC disentangles representations in the text domain and Mel-spectrogram domain into a shared intent space $Z_i$ and domain-specific content spaces $Z_{c_T}, Z_{c_M}$. Intent space aims to encode information related to intent classification from the two domains, while the content feature space aims to encode specific information remaining in each domain.

\subsubsection{Cycle-consist Loss}
To classify medical conditions and use GAN for multi-modal decoupling representations, the premise is that the intent of the text domain and the intent of the Mel-spectrogram domain are theoretically exactly the same. Based on this assumption, as the intent encoders disentangle similar intent representations from the text domain and the Mel-spectrogram domain, we introduce a cyclic-consistency loss function to guide the re-extracted intent representation towards the previously extracted representation after two representation-exchange processes.
The cyclic consistency constraint includes two stages:
\begin{itemize}
\item \textbf{Cross}: Perform intent representation disentanglement and content representation disentanglement in text domain and Mel-spectrogram domain to obtain $ \{ z_{text}^i, z_{text}^c\} $ and $ \{ z_{mel}^i, z_{mel}^c\} $ respectively. Then exchange the intent representations, and generate $ \{ u, v\} $ through generators.
    \begin{equation}
    u = G_{text}(z_{text}^c, z_{mel}^i), \\
    v = G_{mel}(z_{mel}^c, z_{text}^i) 
    \label{eq:uv}
    \end{equation}

\item \textbf{Restore}: Exchange the intent representations again, and restore the original text feature and Mel-spectrogram feature through generators.
    \begin{equation}
    \hat {T} = G(z_u^c, z_v^i), \\
    \hat {M} = G(z_u^i, z_v^c) 
    \end{equation}
    \end{itemize}

After the cross and restore operation, the original text and Mel-spectrogram features are reconstructed. To strengthen this constraint, the cycle-consist loss is defined as follows:
    \begin{equation}
    \mathcal{L}_{cc}(T,M,\hat T,\hat M) = Loss(\hat T, T) + Loss(\hat M, M)
    \label{eq:rs}
    \end{equation}
where $ T $ is the text feature, $ \hat T $ is the reconstructed text feature, $ M $ is the Mel-spectrogram feature and $ \hat M $ is the reconstructed Mel-spectrogram feature. $Loss(\cdot)$ represents the proper criterion that measures the distance between two variables. As shown in Fig. \ref{fig:framework}, $ \hat T $ and $ \hat M $ can be represented by Eq. \ref{eq:tm}.
    \begin{equation}
    \begin{array}{c}
    \hat T = {G_{text}}(E_{text}^{c}(u),E^i_{mel}(v))\\
    \hat M = {G_{mel}}(E_{mel}^{c}(v),E^i_{text}(u))
    \end{array}
    \label{eq:tm}
    \end{equation}
where the construction of $u$ and $v$ is shown in Eq. \ref{eq:uv}.

\subsubsection{Distribution Loss}
In order to enable the intent encoders $ {E_i} $ to learn the same feature from text domain and Mel-spectrogram domain, we let encoders encode the reconstructed features in text and Mel-spectrogram domain to obtain $ \{ E^i_{text}(\hat{T}), E^i_{mel}(\hat{M})\} $ respectively, then use the distribution loss to facilitate approximation between them. 
\begin{equation}
    \mathcal{L}_{distri} = Loss(E^i_{text}(\hat{T}),E^i_{mel}(\hat{M}))
    \label{eq:dis}
\end{equation}
The purpose of Eq. \ref{eq:dis} function is to minimize the distance between them.

\subsubsection{Classification Loss}

After reconstructing the text domain and Mel-spectrogram domain features, we use the intent encoder to disentangle the intent representation from the reconstructed features $ \{ \hat{T},\hat{M}\} $, then fuse and classify the intent features of the two domains, where cross-entropy loss is used as the classification loss. Classification loss shown in Eq. \ref{eq:cls} ensures that the representations disentangled by the intent encoder are intent features related to the classification task.
\begin{equation}
\mathcal{L}_{CE} = H(\hat{y}, y) =  - \sum {y\log (soft\max (\hat{y}))} 
\label{eq:cls}
\end{equation}
where $\hat{y}$ is the network prediction output, $softmax(\cdot)$ presents the prediction output $\hat{y}$ in the form of probability, while $y$ represents the ground-truth label.

\subsubsection{VAE Loss}
To assist the reconstruction of text and Mel-spectrogram features, we use the latent variable space and encourage the representation from the latent variable space to be as close as possible to the prior Gaussian distribution, thus we leverage Kullback Leibler (KL) loss:
\begin{equation}
    \mathcal{L}_{KL} = {\rm E}[{D_{KL}}(({z_c})||Rs)]
\end{equation}
where $ {D_{KL}}(p||q)=\sum\limits_{i = 1}^n {p({x_i})\log \frac{{p({x_i})}}{{q({x_i})}}} $, $Rs$ is a random sample obeys distribution $\mathcal{N}(0,1)$.

\subsubsection{Latent Regression Loss}
We use latent space regression loss to encourage the invertible mapping between the text domain or Mel-spectrogram domain and the latent space. We randomly sample a content latent vector $z$ from Gaussian distribution and try to reconstruct it via the generator $G$:
\begin{equation}
\begin{array}{r}
\mathcal{L}_{lr} = Loss(E_{text}^{c}({G_{text}}(E_{text}^i(T),z_{text}^c), z_{text}^c) + \\
Loss(E_{mel}^c({G_{mel}}(E_{mel}^i(M),z_{mel}^c), z_{mel}^c)
\end{array}
\label{eq:lr}
\end{equation}
where $ z_{text}^c $ is a randomly selected latent vector in text domain, $ z_{mel}^c $ is a randomly selected latent vector in Mel-spectrogram domain.

\subsubsection{Adversarial Loss}
A domain adversarial loss $\mathcal{L}_{adv}$ is also provided, as text or Mel-spectrogram discriminator attempts to discriminate between real feature and generated feature (\textit{i.e.} reconstructed feature) in each domain, and text or Mel-spectrogram generator tries to generate feature closer to reality.

The last three losses are \textit{additional and optional}. Combined with the above analysis and elaboration, it means that the overall objective function of our model is:
\begin{equation}
\begin{array}{c}
\mathop {\min }\limits_{G,E} \mathop {\max }\limits_D \lambda_1\mathcal{L}_{cc} + {\lambda_2}{\mathcal{L}_{distri}} + {\lambda_3}{\mathcal{L}_{CE}}\\
 + ({\lambda_{i}}{\mathcal{L}_{KL}} + \lambda_{ii} \mathcal{L}_{lr} + {\lambda_{iii}}\mathcal{L}_{adv})
\end{array}
\label{eq:total}
\end{equation}

\section{Experiments}

\subsection{Dataset and Data-Preprocessing}
\begin{figure}[htb]
\centering
\includegraphics[width=1.0\columnwidth]{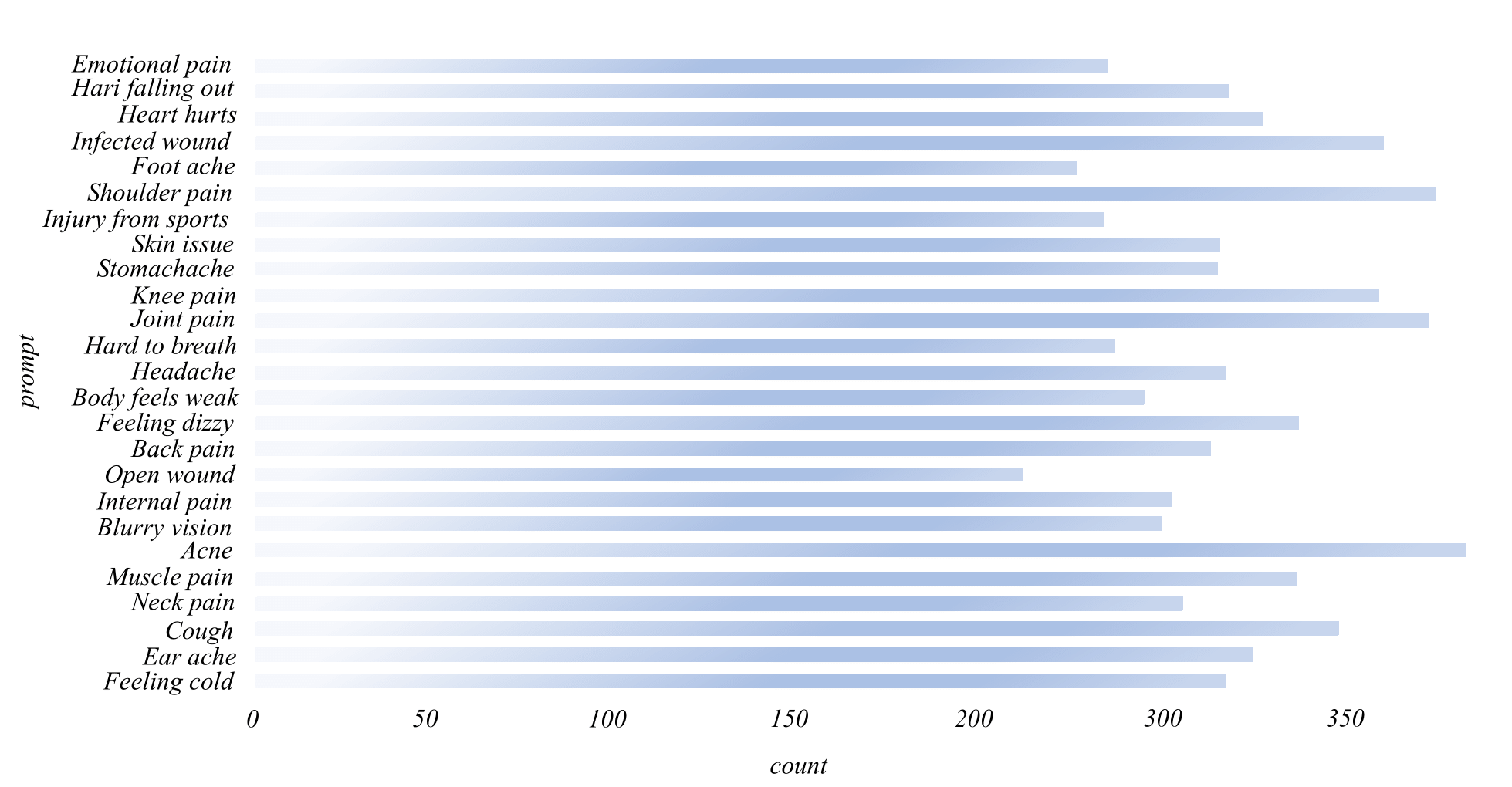}
\caption{Statistics of symptom types in the dataset.}
\label{fig:count}
\end{figure}

We use pre-processed \textit{Medical Speech, Transcription, Intent} dataset to evaluate the classification performance of DRSC. The dataset contains $6,661$ audio segments of varying lengths, including $25$ types of symptom conditions, as shown in Fig. \ref{fig:count}. Due to the low quality of some audio samples in the dataset, we apply a forward-backward digital filter to preserve filter characteristics from the unfiltered signals, thereby reducing their impact on the classification experiments \cite{abdulmohsin2022auto}. These filters perform zero-phase filtering in both the forward and backward directions \cite{gustafsson1996deter}. Short time Fourier transform is conducted to obtain Mel-spectrograms from waveforms. We randomly select $20\%$ of the data from each symptom category as the test set, with approximately $35 \sim 55$ samples per category. To verify the robustness of our classification model, we additionally use the inaccurate text transcriptions obtained from a pre-trained ASR model as text input. 

\subsection{Implementation Details}
A sampling of $16,000$ Hz is used for all the speech audio, $256$-bank Mel-spectrogram is generated through short-time Fourier transform (STFT). The experiment is conducted on a single Tesla V100. We use an Adam Optimizer \cite{kingma2015adam} on a batch of size $16$ with $ {\beta _1} = 0.9 $, $ {\beta _2} = 0.999 $, $ \epsilon = 1 \times {10^{ - 8}} $ and $ lr = 1 \times {10^{ - 4}} $, dropout is applied during training to avoid over-fitting. We experiment with equal weights for each loss component in Eq. \ref{eq:total}.

\begin{figure}[htb]
\centering 
\subfigure[SpeechIC (Mel-Txt-combined)] { \label{fig:a}
\includegraphics[width=1.\columnwidth]{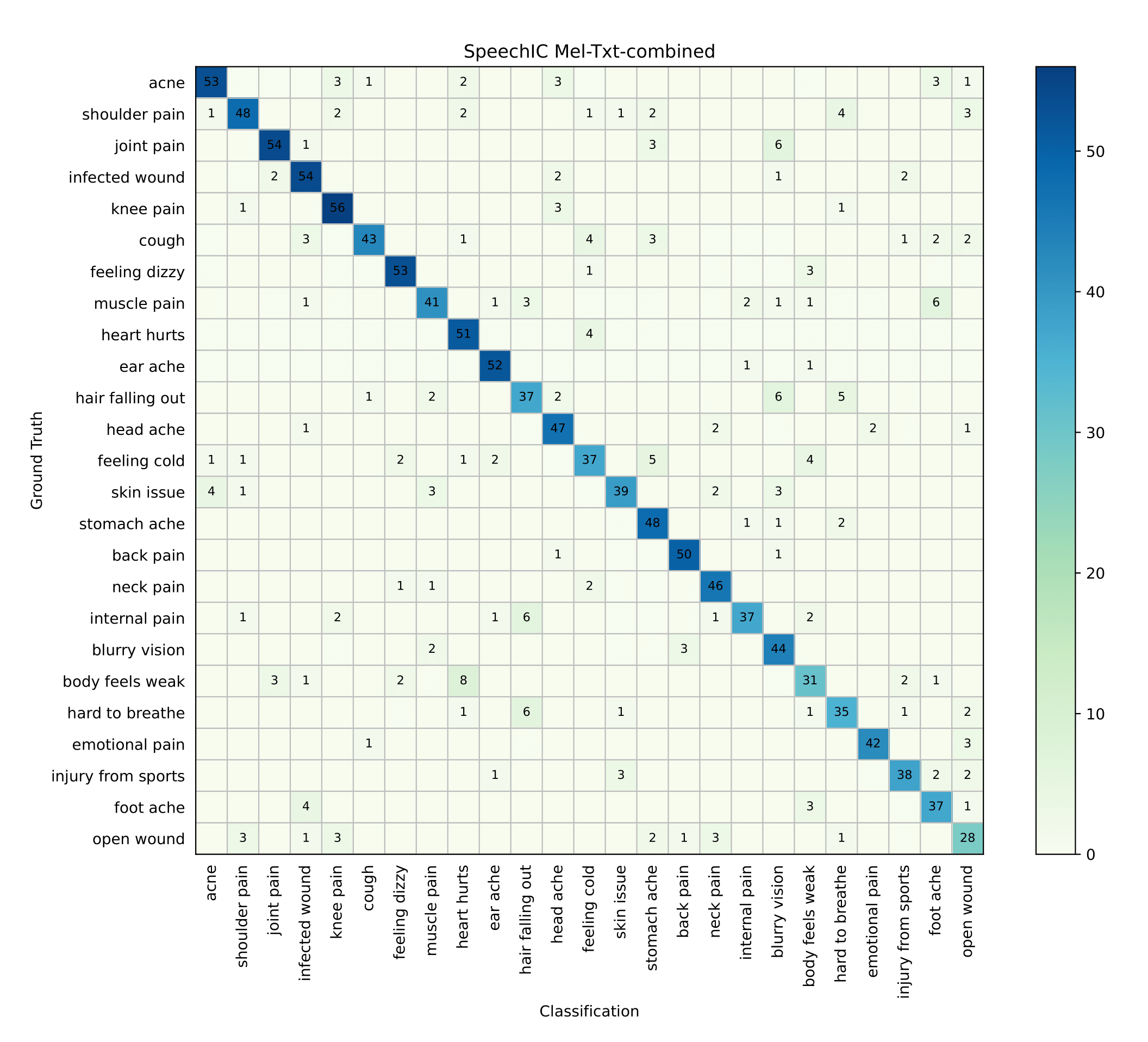}} 
\subfigure[\textbf{DRSC}] { \label{fig:b} 
\includegraphics[width=1.\columnwidth]{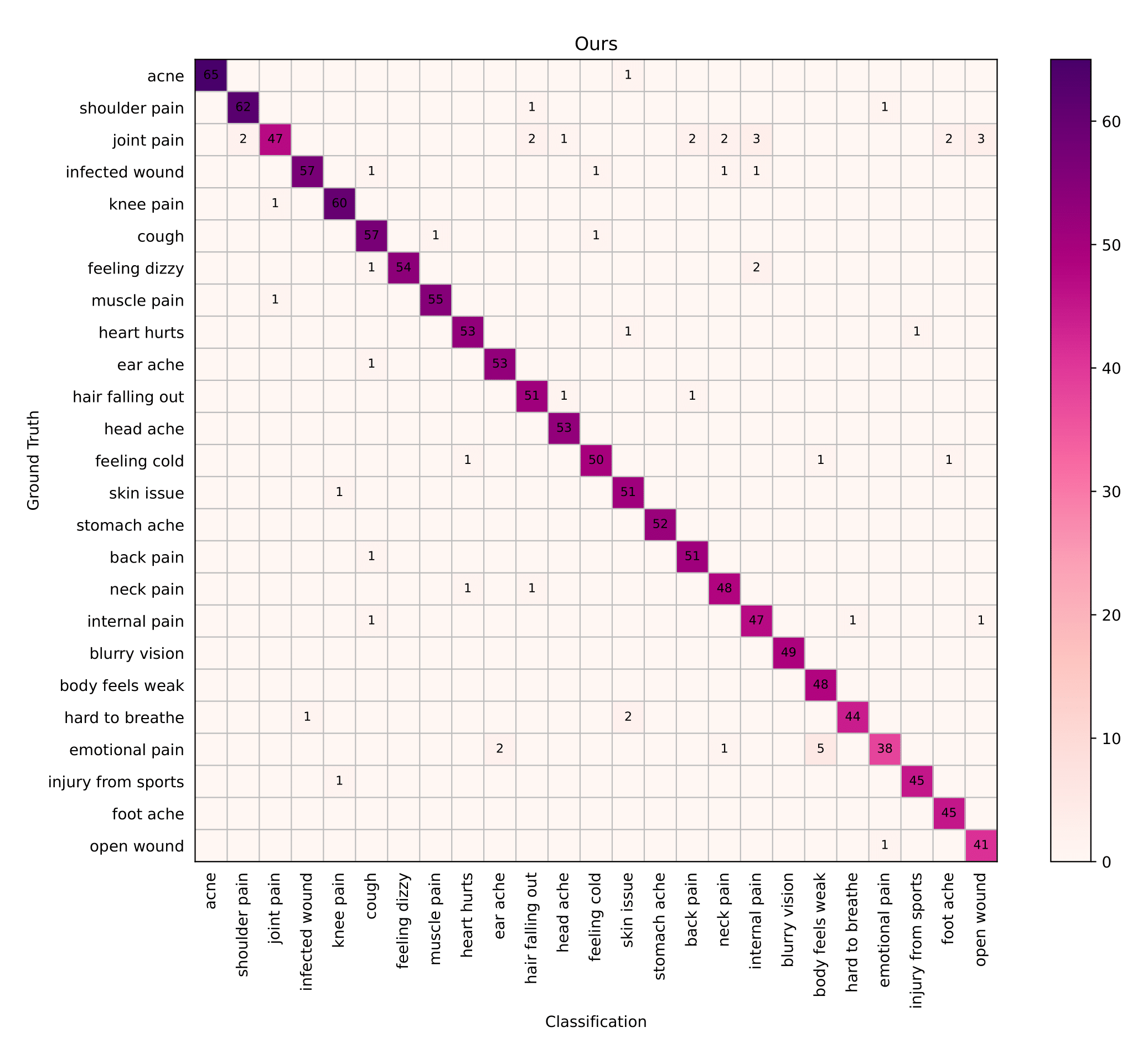}} 
\caption{Confusion matrixes of the baseline method's and proposed method's classification results. SpeechIC uses both Mel-spectrogram and text as input, obtaining better results than using Mel-spectrogram or text only.}
\label{fig:matrix}
\end{figure}

\subsection{Compared Methods}
We use the speech intention classification (SpeechIC), a CNN model proposed by Gu \textit{et al.} \cite{gu2017speechic}, as the baseline model, which takes either Mel-spectrogram, text, or both Mel-spectrogram and text as input. SpeechIC mainly consists of a multi-layer convolutional encoder, a feature fusion layer and a classification prediction layer. The original intention classification model is modified to complete symptoms classification task.
We evaluate the following models to compare the performance of the baseline method and our proposed method.
\\\textbf{SpeechIC Txt-only}: only provide text as input for SpeechIC.
\\\textbf{SpeechIC Mel-only}: only provide Mel-spectrogram as input for SpeechIC.
\\\textbf{SpeechIC Mel-Txt-combined}: provide both Mel-spectrogram and text as input for SpeechIC.
\\\textbf{DRSC}: The model we proposed takes in text and Mel-spectrogram to perform disentangling representation of symptoms intent.

\subsection{Evaluation Results}
Firstly, we select different loss functions for Eq. \ref{eq:rs}, Eq. \ref{eq:dis} and Eq. \ref{eq:lr}, and find out the proper distance criterion. As shown in Table \ref{tab:1}, $L1(\cdot)$ brings the best performance for DRSC, where \textbf{L1 Loss}=$||a, b{||_1}$, \textbf{L2 Loss}=$ ||a, b{||_2}$, \textbf{KL Loss}=${D_{KL}}(a||b)$. Therefore, we choose $L1(\cdot)$ as the distance metric in our loss function for the subsequent experiments.
\begin{table}[htbp]
\centering
\caption{The effect of different distance loss functions.}
\begin{tabular}{cc}
\toprule
Loss Function & Accuracy (\%) \cr
\midrule
\textbf{L1 (\textit{Manhattan distance})} & 95.58 \cr
L2 (\textit{Euclidean distance}) & 94.34 \cr
\textit{cosine distance} & 94.66 \cr
\bottomrule
\end{tabular}
\label{tab:1}
\end{table}

After determining the distance criterion of DRSC, we compare our proposed model with SpeechIC, using the confusion matrix and classification accuracy as evaluation indicators. The accuracy of each model is shown in Table \ref{tab:2}, and the confusion matrix is showcased in Fig. \ref{fig:matrix}. 
\begin{table}[htbp]
\centering
\caption{The classification accuracy of various methods in the classification of medical speech symptoms.}
\begin{tabular}{cc}
\toprule
Method              &Accuracy (\%)        \cr
\midrule
SpeechIC Txt-only              & 67.65            \cr
SpeechIC Mel-only              & 73.04            \cr
SpeechIC Mel-Txt-combined      & 82.47            \cr
\midrule
\textbf{DRSC}                  & 95.58            \cr
\bottomrule
\end{tabular}
\label{tab:2}
\end{table}

The results show that our proposed representation disentanglement model has better accuracy than SpeechIC in medical speech symptoms classification. Our analysis of the confusion matrixes in Fig. \ref{fig:matrix}  finds that our proposed model demonstrates a significant advantage in classification results compared to the baseline model SpeechIC, achieving almost $100\%$ accuracy in the diagnosis of certain symptoms (\textit{e.g.} stomachache, foot ache). However, the classification of certain symptoms like joint pain, is not so satisfactory. 

To verify the effect of the proposed additional optional loss functions, we conduct an ablation study to evaluate the additional loss functions mentioned. It can be easily seen from Table \ref{tab:3} that the additional loss functions bring significant improvement in classification accuracy.
\begin{table}[htb]
\centering
\caption{Explore the effect of the additional loss function.}
\begin{tabular}{cc}
\toprule
Method                                   &Accuracy (\%)           \cr
\midrule
\textit{w/o additional optional loss}      & 81.19                \cr
\midrule
\textbf{DRSC}                              & 95.58                \cr
\bottomrule
\end{tabular}
\label{tab:3}
\end{table}

Finally, considering the difficulty of manually labelling speech text in practical scenarios, we explore the robustness of DRSC, we use the pseudo speech text automatically transcribed via an ASR model as the text input of DRSC, as the ASR model cannot produce a very accurate transcription from speech data, there is a $26\%$ word error rate. However, our results from Table \ref{tab:4} show that despite the use of imprecise transcriptions, DRSC is only $4\%$ lower in prediction accuracy than using accurate text transcriptions. In the case of text-only for SpeechIC, the prediction accuracy is reduced by $9\%$, and the prediction accuracy of SpeechIC using both text and Mel-spectrogram is reduced by $8\%$ in the case of inaccurate transcription of the text, which shows that our model is more robust, making accurate text transcription for medical intent classification is unnecessary.
\begin{table}[ht]
\centering
\caption{Explore the robustness of the model.}
\begin{tabular}{ccc}
\toprule
Method                  &Text input      &Accuracy (\%)            \cr
\midrule
SpeechIC (Txt-only)             &accurate         &67.65      \cr
SpeechIC (Txt-only)             &inaccurate       &58.29            \cr
SpeechIC (Mel-Txt-combined)     &accurate         &82.47      \cr
SpeechIC (Mel-Txt-combined)     &inaccurate       &74.73            \cr
\midrule
\textbf{DRSC}                 &accurate         &95.58            \cr
\textbf{DRSC}                 &inaccurate       &91.43            \cr
\bottomrule
\end{tabular}
\label{tab:4}
\end{table}

\section{Conclusion}
In this paper, we propose a medical classification model named DRSC which extracts intent information from both text and Mel-spectrogram domains for decision. With cross and restore operations, intent representations and content representations are disentangled into the shared intent feature space and domain-specific content spaces, respectively. The intent representations from the text domain and the Mel-spectrogram domain are fused and used for medical classification. Experimental results show that DRSC outperforms existing methods in classification accuracy and owns robustness when facing inaccurate text transcriptions.

\section{Acknowledgement}
This paper is supported by the Key Research and Development Program of Guangdong Province (grant No. 2021B0101400003). Corresponding author is Xulong Zhang (zhangxulong@ieee.org) from Ping An Technology (Shenzhen) Co., Ltd.

\bibliographystyle{IEEEtran.bst}
\bibliography{mybib.bib}

\end{document}